\documentclass[letterpaper, 10 pt, conference]{ieeeconf}  

\IEEEoverridecommandlockouts                              

\usepackage{bm}

\newcommand{\reffig}[1]{Figure \ref{#1}}

\newcommand{\reftab}[1]{Table \ref{#1}}
\newcommand{\refeq}[1]{Equation (\ref{#1})}

\newcommand{\refsec}[1]{Section \ref{#1}}

\newcommand{\lia}[1]{\textcolor{black}{#1}}

\usepackage{float} 
\usepackage{graphicx}
\usepackage{amsmath} 
\usepackage{xcolor}    
\usepackage{cancel} 
\usepackage[normalem]{ulem}

\title{\LARGE \bf
StableTracker: Learning to Stably Track Target via Differentiable Simulation
}

\author{
Fanxing Li, Shengyang Wang, Fangyu Sun,  Shuyu Wu, Dexin Zuo, Yufei Yan, Wenxian Yu, Danping Zou$^\dag$ 
}

\begin{document}

\maketitle
\thispagestyle{empty}
\pagestyle{empty}

\begin{abstract}

\lia{Existing FPV object tracking methods heavily rely on handcrafted modular pipelines, which incur high onboard computation and cumulative errors. While learning-based approaches have mitigated computational delays, most still generate only high-level trajectories (position and yaw). This loose coupling with a separate controller sacrifices precise attitude control; consequently, \textit{even if target is localized precisely}, accurate target estimation does not ensure that the body-fixed camera is consistently oriented toward the target, it still probably degrades and loses target when tracking high-maneuvering target. }
To address these challenges, we present \textbf{StableTracker}, a learning-based control policy that enables quadrotors to robustly follow a moving target from arbitrary viewpoints. The policy is trained using backpropagation-through-time via differentiable simulation, allowing the quadrotor to keep a fixed relative distance while maintaining the target at the center of the visual field in both horizontal and vertical directions, thereby functioning as an autonomous aerial camera. We compare StableTracker against state-of-the-art traditional algorithms and learning baselines. Simulation results demonstrate superior accuracy, stability, and generalization across varying safe distances, trajectories, and target velocities. Furthermore, real-world experiments on a quadrotor with an onboard computer validate the practicality of the proposed approach.

\end{abstract}

\section{Introduction}
Object tracking is a fundamental capability of quadrotors across various applications, including photography and surveillance. It requires quadrotors to follow a moving target while maintaining a stable relative distance and consistent view. Early studies \cite{chen_tracking_2016,qin_perception-aware_2023,ji_elastic_2022} have explored this field extensively and achieved significant progress in deploying tracking policies to real-world scenarios.

\lia{After receiving target position from an external localization module likes YOLO, the quadrotor plans a trajectory and executes it via an outer-loop controller, so as to keep the target both centered in its field of view (FOV) and at a fixed distance.
While the modular design successfully accomplishes this task, its multi-stage architecture—particularly the inclusion of optimization-based planning—introduces significant hardware overhead and cumulative errors. This results in increased control latency, especially on quadrotors with constrained onboard computational resources. Consequently, when the target increases speed or agility, the planner often fails to generate commands in time, leading to target loss.}

To mitigate these limitations, some studies have leveraged reinforcement learning (RL) to train lightweight policies \cite{zhang_back_2024,loquercio_learning_2021} that replace the planning modules, thereby accelerating inference and freeing resources for other onboard programs. However, many learning-based trackers \cite{lu_yopov2-tracker_2025} still formulate the problem as trajectory following: The policy outputs a high-level path, represented as a point mass trajectory, to be tracked by a downstream position controller, while simultaneously adjusting the yaw to keep the camera pointed at the target. However, these high-level controllers lack the ability to directly command attitude (especially pitch and roll), and thus fail to provide explicit control over the target's image-plane centering—typically referred to as vertical bias in the FPV perspective. \textit{\textbf{Even with a precise target location}}, when the target rapidly accelerates or decelerates, the quadrotor must pitch down/up to adjust its speed, which can drive the target out of the FOV. 
Therefore, current control method is not theoretically capable to track a maneuvering object. 


\begin{figure}[t]
    \centering
    \includegraphics[width=1.0\linewidth]{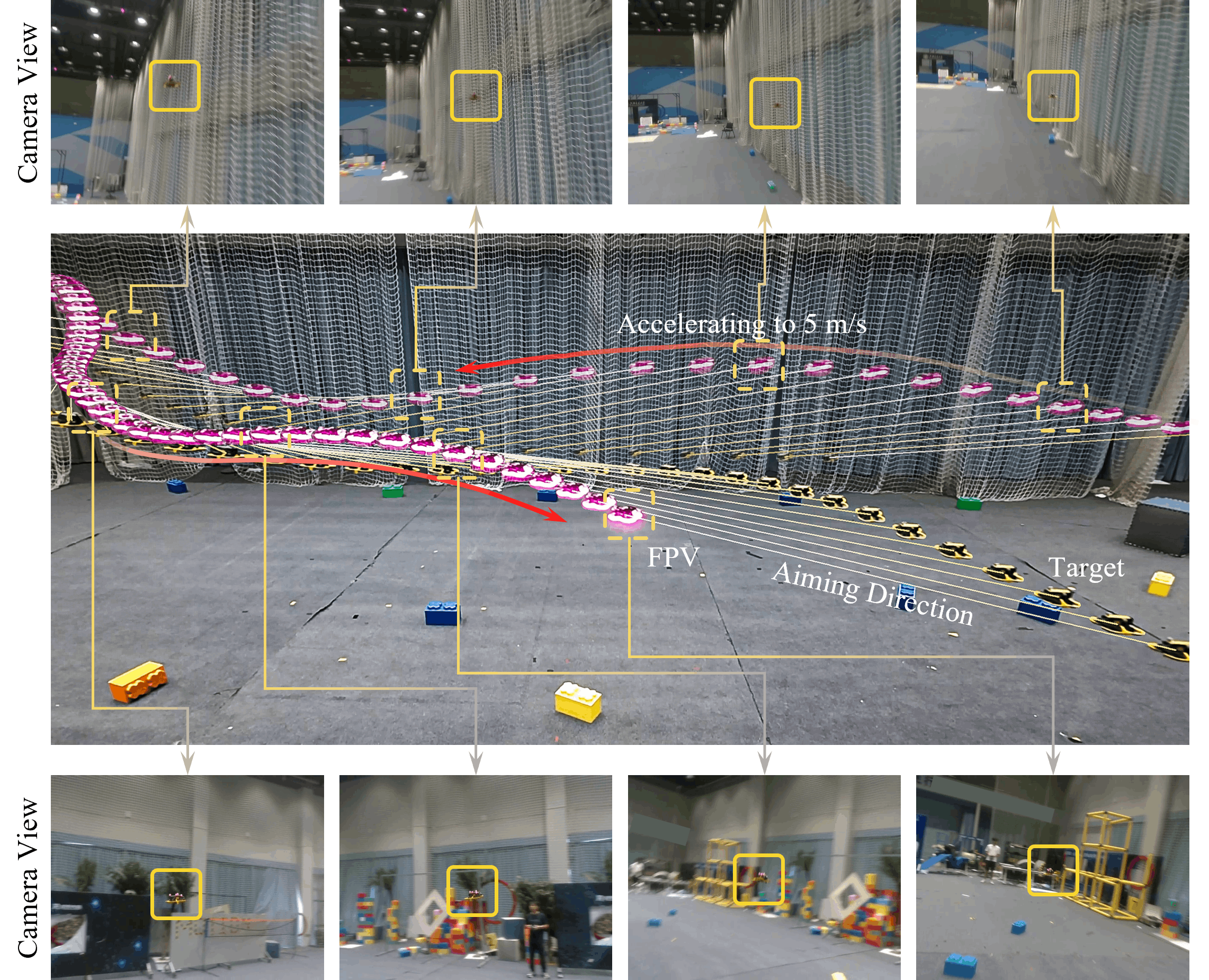}
    \caption{The schematic of real-world experiment. Our FPV quadrotor autonomously tracks a maneuvering DJI quadrotor, centering the target and keeping fixed distance.}
    \label{fig:exp_real_world}
\end{figure}

Utilizing low-level commands like collective thrust and bodyrates (CTBR) enables to control the quadrotol much tightly,  as low-level control grants full, dynamic control of the vehicle’s orientation. Recently, first-order gradient RL facilitated by differentiable simulation has shown remarkable effectiveness in training quadrotor policies, for instance, in aggressive flight in cluttered environments \cite{zhang_back_2024}. Even in stabilization tasks such as trajectory tracking \cite{li2025abpt}, compared to standard RL algorithms such as PPO \cite{schulman2017proximal}, differentiable physics has yielded policies with significantly smoother control outputs,\lia{ especially low-level control.} 

Motivated by these observations, we propose \textbf{StableTracker}, a learning-based policy designed to \textbf{stably} track the moving target while maintaining a constant relative distance and centering the target within the field of view. \lia{It is the first policy in tracking task with low-level commands, which could precisely adjust the attitude of quadrotors for monitoring target.} By leveraging differentiable simulation during training, the proposed policy overcomes {the poor generalization and sub-optimal convergence} of traditional RL approaches.  We compare StableTracker with advanced planning-based baselines, and the results demonstrate that our method achieves superior stability and maintains the target at the center of FOV more precisely. {It also exhibits strong generalization and robustness to variations in object velocity, trajectory patterns, and safe distance.} The contributions of this work are summarized as follows:
\begin{itemize}
    \item To the best of our knowledge, we first demonstrate the capacity of differentiable simulation to train the policy in varying external interaction environment.
    
    \item We propose StableTracker, a lightweight and deployable onboard policy that enables quadrotors to stably track the target while maintaining a constant relative distance.
    
    \item \lia{ The policy directly outputs low-level commands to enhance maneuverability, allowing rigidly-mounted cameras to track targets much more accurately. 
    }
\end{itemize}

\section{Related Work}

FPV object tracking tasks can be categorized into several subfields, including target capture 
\cite{wang_image-based_2023}, 
visibility-aware flight, and autonomous landing \cite{gao_adaptive_2024}. Among these, visibility-aware flight places the greatest emphasis on maintaining stable target visibility and a consistent relative distance.  

Traditional studies on visibility-aware flight are mostly based on planning-based methods. Han \cite{han_fast-tracker_2021} first achieved agile tracking in unseen cluttered environments relying on visual perception, and its subsequent work \cite{pan_fast-tracker_2021} improved the stability by equipping quadrotor with a gimbal. Chen \cite{chen_tracking_2016} formulated a short-horizon optimization problem that minimizes the cost defined by the target’s distance. Wang \cite{wang_visibility-aware_2021} simultaneously optimized yaw and position control to improve target visibility in cluttered environments. Ji \cite{ji_elastic_2022} introduced an adaptive relative-distance factor to relax hard visibility constraints and enhance safety. Building upon this, Gao \cite{gao_adaptive_2024} extended the framework to enable aerial robots to perch on moving platforms. More recently, Qin \cite{qin_perception-aware_2023} increased target visibility and distance while executing aggressive maneuvers through a novel underactuation compensation scheme. However, to achieve real-time inference, these methods rely on high-performance onboard computers; otherwise, they suffer from severe control latency and cumulative error. 

To address this limitation, researchers have turned to RL for policy learning. Sampedro \cite{sampedro_image-based_2018} demonstrated a drone capable of tracking a ground vehicle using a downward-facing camera, though this approach remains constrained by the limited maneuverability of ground targets. A series of works 
\cite{xi_anti-distractor_2022, zhao_hierarchical_2021} 
employed backbones to map images directly to actions, but these works were mainly validated on simulated datasets due to the photo-realistic gap.
Dionigi \cite{dionigi_d-vat_2024} firstly validated its learning algorithm in hardware-in-loop simulation. Nonetheless, directly image-to-action still suffers from sim-to-real gap, making it difficult to deploy in real world. 
An extraordinary work YOPO \cite{lu_yopov2-tracker_2025} replaces a single time-consuming planning module and generate waypoints with learned components, firstly achieving high-speed tracking as high as 7m/s. 
However, most of works including YOPO plan trajectories and use a following outer-loop controller to track, unable to precisely control altitude, making it difficult to keep target in FOV vertically. Moreover, this approach may occasionally generate dynamically-infeasible trajectories, as it ignores the inherent physical principles of quadrotors, which is why traditional pipelines include a time-reallocation \cite{zhou2020ego} module after planning the waypoints.  


\section{Methodology}
\lia{
Based on the current state of research, the most practical and feasible pipeline to track a recognized target is a policy directly outputs low-level collective thrust and bodyrates (CTBR) commands, for stronger maneuverability and much precise attitude control. 
}

\subsection{Problem Formulation}

\begin{figure}[h]
    \centering
    \includegraphics[width=0.7\linewidth]{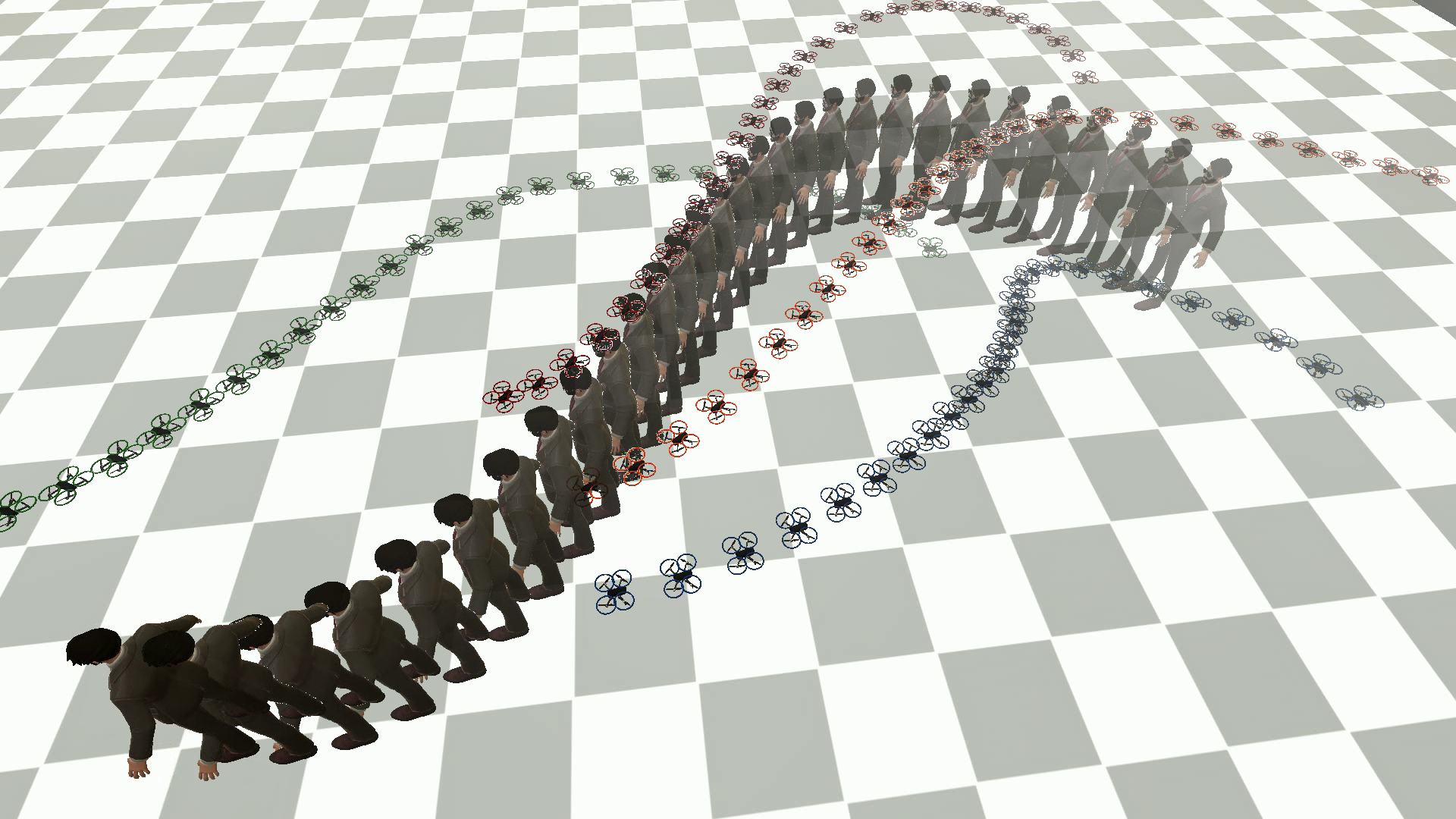}
    \caption{Four quadrotors start from four orthogonal positions to fairly evaluate generalization capability. They then track a moving human target, attempting to keep the target centered within the FOV while maintaining a safe relative distance. (Rendered by VisFly \cite{li2024visfly})}
    \label{fig:objTrack}
\end{figure}

A perception-aware stable tracking task can be defined as follows: as illustrated in \reffig{fig:objTrack}, the quadrotor initially hovers at a safe relative position while facing the target human. As the target moves, the quadrotor is required to keep the target centered within its FOV while simultaneously maintaining a safe distance, even when the target executes abrupt maneuvers. A forward-facing camera is rigidly mounted at the head of the quadrotor. \lia{
As we focus on algorithm of training a tracking policy, we assume the position of target is known instead of object detection and localization.
}

\lia{
We formulate the tracking problem as a Markov process. Given the state of quadrotor $s_k=\{\mathbf{p}_W, \mathbf{q}_W, \mathbf{v}_W, \mathbf{\Omega}\}\in\mathcal{S}$ comprised of position, orientation, linear velocity and angular velocity in the world frame, and the state of target $s^t_k=\{\mathbf{p}^t_W, \mathbf{v}^t_W\}\in\mathcal{S}^t$ comprised of position and linear velocity in the world frame, at each time step, the policy  $\pi_\theta(\cdot | o)$ obtains observation $o_k\in\mathcal{O}$ and infers the action $a_k$ =$ \left [f^d, \mathbf{\Omega}^d\right]\in\mathcal{A}$ including collective thrust and bodyrates. The quadrotor executes the action, receives a reward $r_k=R(s_k, a_k)$ and transits to the next state $s_{i+1}$, where $R(\cdot)$ denotes the reward function.}

\lia{
Considering real-world implementation, the quadrotor receives a noisy observation $o_k$ from onboard sensors and estimators. Besides, we transform the states in the heading frame, since projecting states onto this frame substantially reduces the variance of observations, thereby facilitating training. This observation $o=\{\mathbf{v}_H,\mathbf{q}_W,\mathbf{\Omega},\mathbf{p}_H,\mathbf{v}_H\}$ includes the FPV state: the linear velocity expressed $\mathbf{v}_H$ in the heading frame, orientation represented as a quaternion in the world frame, angular velocity in the body frame, as well as the target state in the heading frame: the estimated linear velocity $\mathbf{v}_H^t$ and position $\mathbf{p}_H^t$, which are
}
\begin{equation}
\begin{aligned}
        \mathbf{v}_H^t = & \mathbf{R}_{WH}(\mathbf{v}^t_W - \mathbf{v}_W) \\
    \mathbf{p}_H^t = & \mathbf{R}_{WH}(\mathbf{p}^t_W - \mathbf{p}_W) \\
        \mathbf{v}_H = & \mathbf{R}_{WH}\mathbf{v}_W \\
\end{aligned}
\end{equation}
where $\mathbf{R}_{WH}$ denotes the transformation matrix from the world frame to the heading frame.  The heading frame is obtained by rotating the world frame by the yaw angle $\psi$, and then $\mathbf{R}_{WH}$ could be expressed as:
\begin{equation}
\mathbf{R}_{WH} = 
    \begin{bmatrix}
    \cos\psi & \sin\psi & 0 \\
    -\sin\psi & \cos\psi & 0 \\
    0 & 0 & 1
    \end{bmatrix}
\end{equation}


\subsection{Why Learning via Differentiable Simulation}
\label{sec:bptt}

Model-free reinforcement learning methods can generally be categorized into two classes: value iteration and policy iteration. The most widely adopted framework in both categories is the actor–critic architecture. In this paradigm, the critic $Q_{\phi}(s, a)$ estimates the long-term return using experience tuples $\{o_k, a_k, r_k, d_k, o_{k+1}\}$, consisting of observations, actions, rewards, termination indicators, and subsequent observations, where actions are generated by the actor. The actor $\pi_\theta$ is then updated to maximize the long-term return predicted by the critic.  




The gradient estimated by model-free RL methods to update the actor—also referred to as the zero-order gradient—is an unbiased approximation by sampling of the true first-order gradient, although with non-negligible variance. This approach has proven effective in early RL benchmarks such as 
Go and video games. 
Although these algorithms are capable of training policy adequate to use, they always suffer from generalization, which makes it fragile for deployment. 
Besides, for fully differentiable processes such as robotics, and in particular quadrotor dynamics, model-free RL neglects the inherent differentiability of the underlying physics. 

To overcome this limitation, backpropagation through time (BPTT) \cite{mozer2013focused} leverages first-order gradients within differentiable simulations, enabling precise computation of the actor’s gradient.  Compared with zero-order estimation, BPTT offers lower gradient variance, faster convergence, and improved sample efficiency, making it particularly suitable for continuous-control tasks in robotics. The accumulative reward $\mathcal{J}_\theta$ is expressed in \refeq{eq:first_grad}:
\begin{equation}
\label{eq:first_grad}
\mathcal{J}_\theta = \left( \sum_{k=0}^{N-1} \gamma^k {R(s_{k},a_{k})} \right), a_k\sim\pi_\theta(\cdot|o_k)
\end{equation}
where $N$ represents the length of horizon and $\gamma$ denotes the discount factor for emphasizing recent awards. The first-order gradient to optimize actor can be computed through backward chain propagation, expressed as
\begin{equation}
\begin{aligned}
    &\nabla_\theta \mathcal{J}_\theta = \\
    &\sum_{k=0}^{N-1} \gamma^k \Bigg [  \sum_{i=1}^k \frac{\partial R(s_{k},a_{k})}{\partial s_k} \prod_{j=i}^k \left( \frac{\partial s_j}{\partial s_{j-1}} \right) \frac{\partial s_i}{\partial a_i}\frac{\partial a_i}{\partial \theta}   \\
    &+  \frac{\partial R(s_k,a_k)}{\partial a_k} \frac{\partial a_k}{\partial \theta } \Bigg ]
\end{aligned}
\end{equation}
For quadrotors, the state transition process is defined as $\mathcal{S}\times\mathcal{A}\rightarrow\mathcal{S}$, where the dynamics are governed by the differential equation $\dot{s}=F(s,a)$. In practical deployment, this continuous process is discretized into a time-stepped evolution of the system, expressed as $s_{k+1} = s_k + F(s_k, a_k)\,dt$.
\lia{
The dynamics function $F$ can be explicitly represented as the set of equations in \refeq{eq:dynamics}:
}
\begin{equation}
\small
\label{eq:dynamics}
\begin{aligned}
    &\dot{\mathbf{p}}_W = {\mathbf{v}}_W \\
    &\dot{\mathbf{v}}_W = \frac{1}{m} \mathbf{R}_{WB}(\mathbf{f}  + \mathbf{d}) + \mathbf{g} \\
    &\dot{\mathbf{q}} = \frac{1}{2} \mathbf{q} \otimes \mathbf{\Omega} \\
    &\dot{\mathbf{\Omega}} = \mathbf{J}^{-1} \left (\boldsymbol{\eta} - \mathbf{\Omega} \times \bm{J} \mathbf{\Omega} \right )
\end{aligned}
\end{equation}
Here, the quaternion–vector product is denoted by $\otimes$, and $\mathbf{R}_{WB}$ represents the rotation matrix from the body frame to the world frame. The collective rotor thrust, aerodynamic drag, and  moments are denoted by $\mathbf{f}$, $\mathbf{d}$, and $\boldsymbol{\eta}$, respectively. The quadrotor inertia is modeled with a diagonal matrix $\mathbf{J}$, and the vehicle mass is denoted by $m$. Noting that here $\mathbf{f}$ could be expressed as $[0,0,f]$, since the motors' force is aligned with z-axis. The terms $\mathbf{f}$ and $\boldsymbol{\eta}$ represent the actual control computed by the differentiable PD controller from action.
Further details regarding the complete dynamics model and its differentiable formulation can be found in the VisFly \cite{li2024visfly}. The partial derivatives $\partial s_{k+1}/\partial s_{k}$ and $\partial s_{k}/\partial a_{k}$ can be directly inferred from the system of equations in \refeq{eq:dynamics}. At this point, the overall differentiable process is fully established, enabling the application of BPTT.

\begin{figure*}[h]
    \centering
    \includegraphics[width=1.0\linewidth]{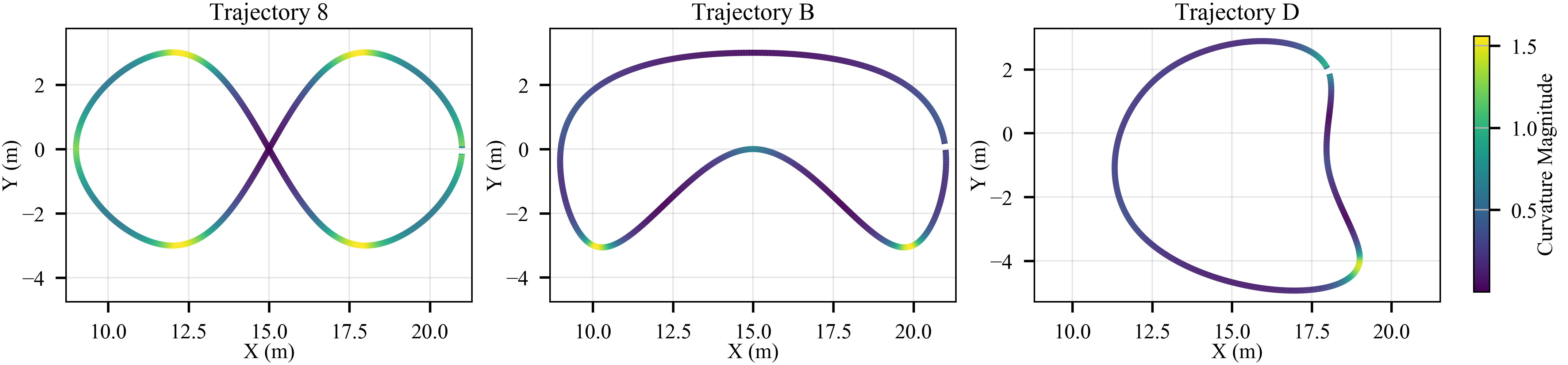}
    \caption{Unseen trajectories of moving targets used in the evaluation experiments, named according to their geometric resemblance. The scale and curvature variance gradually increase from trajectory~8 to trajectory~D. 
    }
    \label{fig:traj}
\end{figure*}

\subsection{Reward Structure}
The reward function for backpropagation-through-time (BPTT) must be fully differentiable\cite{li2025abpt}.
We design a differentiable reward function with three components: a perception-aware reward, a constant-distance reward, and a stability reward, as defined in \refeq{eq:reward_func}:
\begin{equation}
    \label{eq:reward_func}
    R(s_k,a_k)=c_0R^{per}(s_k)+c_1R^{dis}(s_k)+R^{sta}(s_k,a_k)+c
\end{equation}
where $c$ is a constant baseline term for alive, and $c_0, c_1$ are weighting factors for the perception and distance components, respectively.

Given the onboard camera is mounted in the heading direction, the camera $x$-axis is aligned with the quadrotor body-frame $x$-axis $\mathbf{x}_B$. The projection of the relative position vector onto the camera $x$-axis therefore reflects how well the quadrotor is oriented toward the target. Accordingly, as given in \refeq{eq:r_per}, this reward is used to encourage the quadrotor to face the target. 
\begin{equation}
    \label{eq:r_per}
    R^{per}(s_k) = 
    \frac {\mathbf{x}_B \cdot (\mathbf{p}^t_W-\mathbf{p}_W) }
    {\left \| \mathbf{p}^t_W-\mathbf{p}_W \right \|_2 }
\end{equation}

The desired safe distance between the quadrotor and the target is a constant $d$.The distance reward, defined as \refeq{eq:r_dis}, penalizes deviations from $d$. Here we set $d$ to be 3 m in our experiments. 
\begin{equation}
    \label{eq:r_dis}
    R^{dis}(s_k) = -\left | \left \| \mathbf{p}^t_W-\mathbf{p}_W \right \| _2-d \right |
\end{equation}

The stability reward in \refeq{eq:r_sta} penalizes high-frequency fluctuations and non-smooth actions. For a real-world tracking quadrotor equipped with a three-axis gimbal, a stable constant bias is far more acceptable than unpredictable jitter. The first two terms penalize the magnitude of linear and angular velocities, while the last term suppresses sharp variations between consecutive control commands. 
\begin{equation}
    \label{eq:r_sta}
    R^{sta}(s_k,a_k) =  
    -c_2\left \| \mathbf{v}_W \right \| _2
    -c_3\left \| \mathbf{\Omega}_W \right \| _2
    -c_4\left \| a_{k+1}-a_k \right \| _2
\end{equation}

\subsection{Domain Randomization}
To enhance the generalization capability of the learned policy, we introduce randomization in the trajectories and velocities of the moving target. The simulation executes multiple scenes in parallel and updates these randomized settings after a fixed number of timesteps during training. 
Among all dynamic parameters, only the aerodynamic drag coefficients are randomized, as other parameters can be accurately identified.
It should be noted that while randomization improves generalization, it also increases the difficulty of policy convergence.  
This choice is motivated by the fact that drag is strongly environment-dependent and difficult to measure precisely, whereas parameters such as mass and gravity can be reliably modeled. 
The quadrotors are initialized at random positions 1-5 m from the target. To avoid losing visual contact at initialization,  they start with the same initial velocity as the target and orientation directed toward the target.

\section{Experiments}
To evaluate the performance of the policy trained with differentiable simulation, we conducted a series of experiments in both simulation and real-world environments. \lia{We selected Elastic Tracker \cite{ji_elastic_2022} as the baseline of trajectory-control method, which is the widely-recognized as well as open-source algorithm focusing on perception-aware flight among all state-of-the-art methods.} In addition, we trained policies using PPO \cite{schulman2017proximal} and SAC \cite{haarnoja2018soft}, as these algorithms are widely recognized in robotics for their effectiveness in policy learning. Three representative trajectories, illustrated in \reffig{fig:traj}, were employed to ensure a fair comparison across all approaches. The tracking results of all policies are visualized and included in the supplementary video.

\subsection{Policies Training}
\label{sec:training}
We deploy BPTT by ourselves based on SHAC \cite{xu2022accelerated}, an enhanced first-order gradient algorithm. To ensure fairness, all algorithms are implemented with the same stochastic policy network architecture, consisting of a simple three-layer multilayer perceptron with 512 units per layer. For fair comparison, we align algorithms' hyperparameters as closely as possible, and thoroughly explore the optimal specific parameters.  
We simultaneously run 32 parallel environments including various trajectories and 256 agents. The trajectories are refreshed every 3 million timesteps.

\begin{figure}[h]
    \centering
    \includegraphics[width=1.0\linewidth]{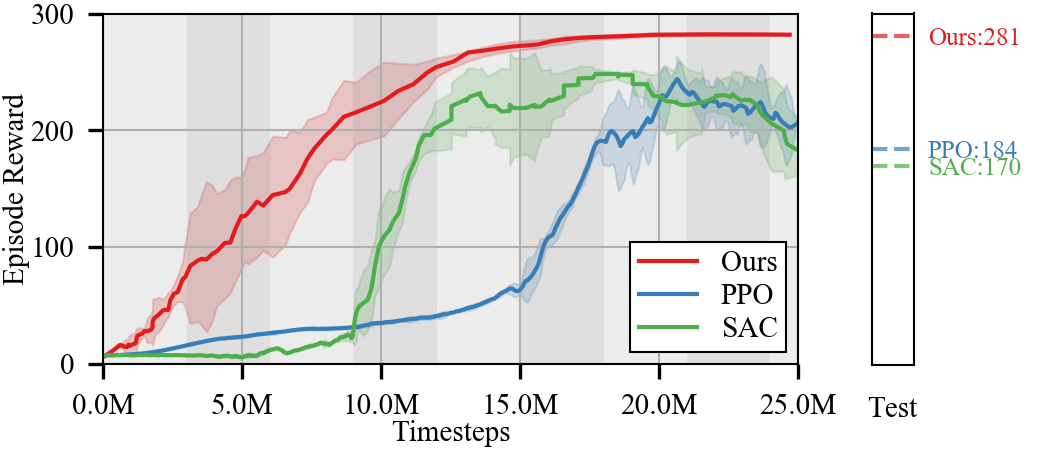}
    \caption{Cumulative rewards curves against timestep of PPO, SAC, and ours. The right axis denotes the reward in test mode. All the scenes in environments will be refreshed when the gray scale changes. All experiments were conducted on a desktop equipped with an NVIDIA RTX 4090 and an Intel Core i9-13900K, with training time of approximately two hours.}
    \label{fig:train_curve}
\end{figure}

As shown in \reffig{fig:train_curve}, ours achieves more stable convergence and reaches higher rewards faster than both PPO and SAC. This result is consistent with the fact that first-order gradients provide more accurate optimization directions than zero-order approximations. In addition, we observe both PPO and SAC overfit the trajectories since the rewards in test mode are lower than expected. This phenomenon  could be observed by the sudden drop when it comes to new scene in \reffig{fig:train_curve}. In contrast, ours remains robust and converges smoothly, thereby demonstrating the superior generalization capability of first-order gradient methods. Since both model-free algorithms converge to similar ultimate reward, we only choose PPO for comparison in the following experiments.

\subsection{Metrics Definition and Experimental Settings}
We define the evaluation metrics for quantifying how well the quadrotor tracks the target as in \refeq{eq:metric}. Given a camera resolution of width $w$ and height $h$, the perception-aware quality is  measured by the normalized distance on the image plane between the projected target pixel and the principal point. 
Because of dynamic constraints and high-level command limits, most previous methods prioritize maintaining horizontal rather than vertical centering, whereas our algorithm achieves a more balanced treatment of errors in both dimensions.
To provide a more comprehensive analysis, perception metric is computed along both horizontal and vertical dimensions. The distance error is defined as the deviation between the actual distance to the target and the desired safe distance $d$:  

\begin{equation}
\label{eq:metric}
\begin{aligned}
    e_H &= 2\frac{p^t_{H}-w/2}{w}, \\
    e_V &= 2\frac{p^t_{V}-h/2}{h}, \\
    e_d &= \left \| \mathbf{p}^t_W-\mathbf{p}_W \right \| _2 - d,
\end{aligned}
\end{equation}
where $p^t_{H}$ and $p^t_{V}$ denote the horizontal and vertical pixel coordinates of the target in the image plane, respectively.

\begin{figure*}[h]
    \centering
    \includegraphics[width=1.0\linewidth]{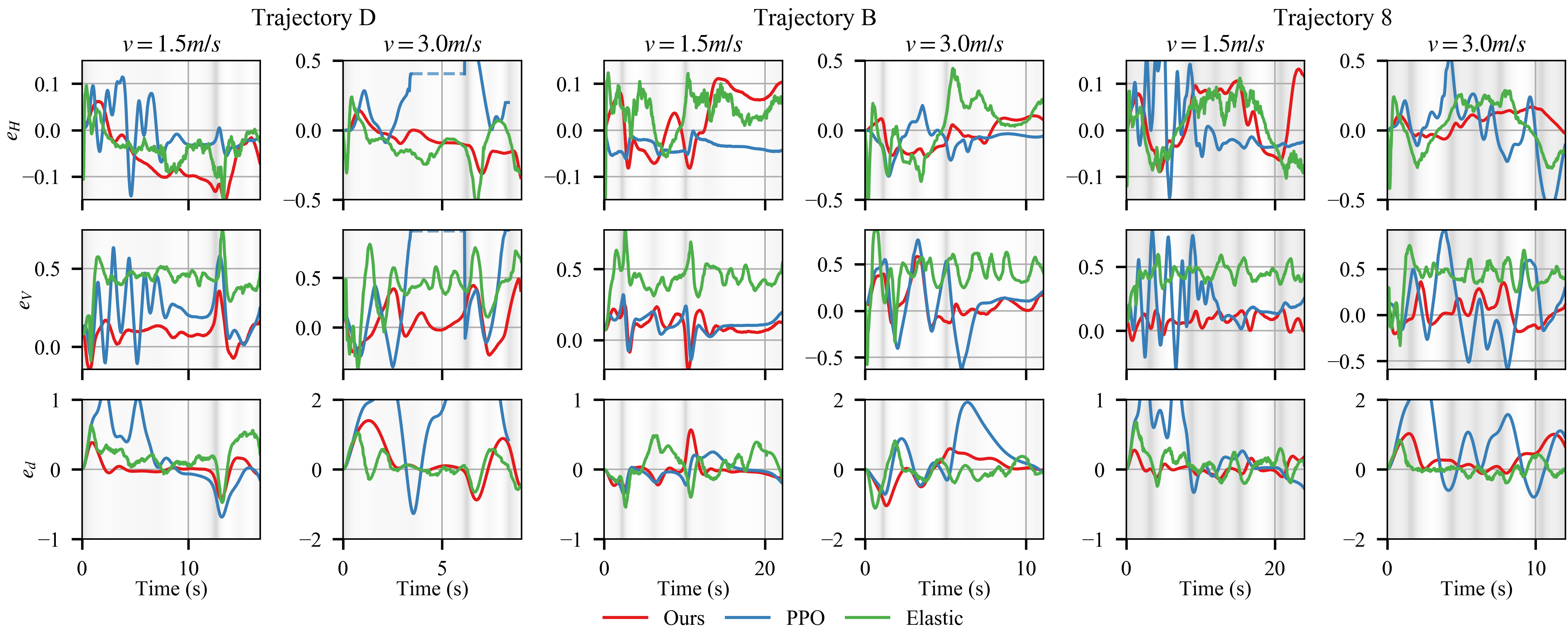}
    \caption{Error curves of different policies over time while tracking a target moving along three trajectories with velocities of 1.5\,m/s and 3.0\,m/s. The \textbf{left}, \textbf{middle}, and \textbf{right} column pairs correspond to Trajectories~D, B, and~8, respectively. The shaded gray background denotes the instantaneous curvature of the trajectory, while the dashed line indicates moments when the quadrotor loses visual contact with the target.}
    \label{fig:exp_vel}
\end{figure*}

We conduct comprehensive experiments to compare the performance of our method against baseline methods. The evaluations are carried out across different velocities, trajectories, and tracking perspectives. To emulate typical human motion, the target is commanded to move at constant speeds ranging from 0.5\,m/s to 3.0\,m/s, covering behaviors from casual walking to running. All quadrotors are initialized with the same velocity as the target and oriented toward it, positioned at a predefined distance $d$. 

Regarding trajectories, as shown in \reffig{fig:traj}, the evaluation consists of three unseen  circular paths including the commonly used figure-eight trajectory, with increasing difficulty as the target motion becomes more complex. To further assess generalization, four quadrotors are simultaneously deployed from four orthogonal positions relative to the target (front, left, right, and rear), as illustrated in \reffig{fig:objTrack}. The detailed experimental settings are summarized in \reftab{tab:exp_config}.

\begin{table}[htbp]
\centering
\caption{Experimental Configuration}
\label{tab:exp_config}
\begin{tabular}{l|l}
\hline
\textbf{Parameter} & \textbf{Value} \\
\hline\hline
Velocity ($m/s$) & 0.5,1.0,1.5,2.0,2.5,3.0 \\
Trajectory & D, B, 8 \\
Initial Position (clock angle) & 0°, 90°, 180°, 270° \\
\hline
\end{tabular}
\end{table}

\subsection{Sudden Target Acceleration and Deceleration}
\label{sec:acc_dec}
{When required to maintain a fixed safe distance, the quadrotor must pitch down and accelerate to follow a target undergoing acceleration, which causes the camera to tilt downward. Conversely, it must pitch up to provide a braking force when the target decelerates, which tilts the camera upward. As a result, the target drifts away from the image center, and under extreme acceleration may even move out of the FOV. This inherent limitation arises naturally with high-level waypoints commands, as they cannot efficiently regulate the pitch angle. }
In contrast, our learned control policy directly outputs low-level CTBR commands, providing greater maneuverability to handle such cases.

\begin{figure} 
    \centering
    \includegraphics[width=1.0\linewidth]{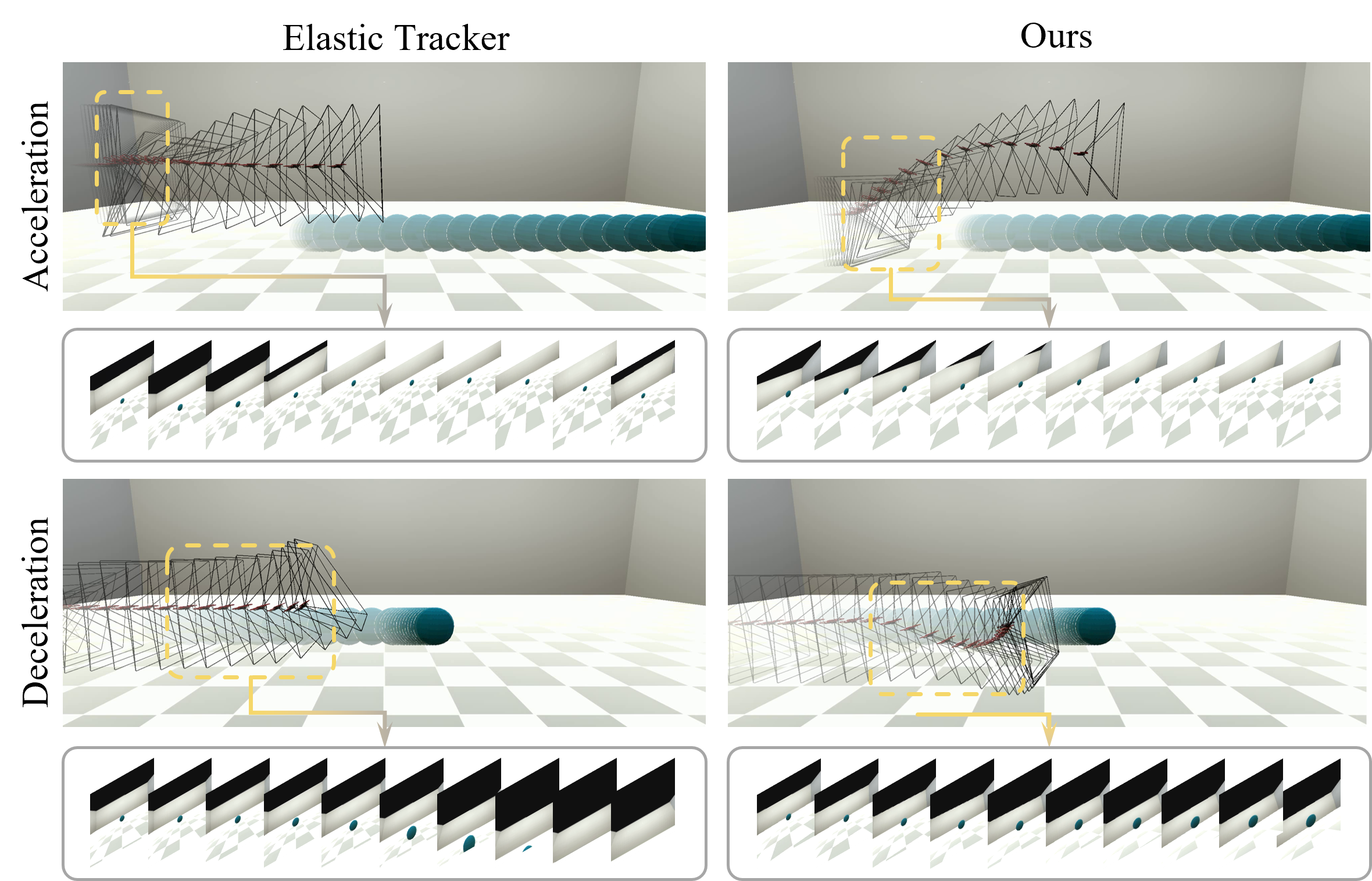}
    \caption{Tracking performance of our method compared with Elastic under rapid acceleration and deceleration. In the test, the target accelerates from 0.5 m/s to 4.0 m/s within 0.2 m, and conversely decelerates from 4.0 m/s to 0.5 m/s. The bottom row of each subfigure presents the sequential camera observations within the highlighted yellow box.}
    \label{fig:exp_acc_dec}
\end{figure}

Interestingly, our policy enables the quadrotor to learn a novel maneuver. When the target suddenly brakes, the quadrotor pitches up to generate the required decelerating force while simultaneously descending. This descent compensates for the upward camera tilt, keeping the target centered in the FOV; the opposite adjustment occurs (ascent) when the target accelerates. We refer to these behaviors as \emph{coordinated descent and ascent}. As shown in \reffig{fig:exp_acc_dec}, because it tracks the target from a higher position, Elastic Tracker exhibits fluctuations during acceleration and loses the target during deceleration. In contrast, our policy consistently maintains target centering under both extreme scenarios. Additionally,  a repeated deceleration–acceleration experiment is provided in the supplementary video.

\subsection{Trajectory-Level Tracking Performance}

For aerial photography tasks, \textbf{stability} is more critical than absolute accuracy, as constant bias can be effectively estimated and compensated by a camera gimbal. 
To evaluate stability at the trajectory level, we plot the tracking errors over the entire target trajectory, as illustrated in \reffig{fig:exp_vel}.
To compare the performance of our policy against the baselines, we consider two typical target velocities: a normal walking speed of 1.5\,m/s and a normal running speed of 3.0\,m/s. The quadrotor is initialized at the front side of the target, representing a common perspective in practice, since humans usually move forward while facing the camera. As shown in \reffig{fig:exp_vel}, the curves across all algorithms share several common characteristics:

\begin{itemize}
    \item Continuous variations in target motion make perfectly precise tracking impossible, resulting in fluctuations of different magnitudes across all methods.
    \item Errors increase sharply at turning points, where the target trajectory deviates substantially from the prediction.
    \item Among the influencing factors, \textbf{trajectory curvature} plays the most significant role rather than speed in degrading tracking stability, as the future path of the target becomes increasingly difficult to predict. 
\end{itemize}
Beyond these common features, each algorithm also exhibits distinctive behaviors, which are analyzed in detail below.

\textbf{PPO}: PPO exhibits pronounced oscillations at the beginning of tracking, as the initial relative state may fall outside its randomization domain. At lower target velocities, despite the initial instability, the policy can gradually converge and eventually achieve performance competitive with ours. However, at higher velocities the convergence process becomes significantly slower, and in extreme cases the vertical error $e_V$ even exceeds~1, making PPO the only algorithm that fails to track target. Such a failure is observed at 3\,m/s along Trajectory~D, as shown in \reffig{fig:exp_vel}. Overall, these results are consistent with the poor generalization ability of PPO discussed in \refsec{sec:bptt} and \refsec{sec:training}.

\textbf{Elastic Tracker (Elastic)}: For fair comparison, we use Elastic's built-in simulator and controller, then render the trajectory and motions by VisFly. Elastic exhibits a large horizontal error $e_H$ at the beginning of tracking, as it requires time to re-plan a privileged trajectory. Although the overall errors remain within a tolerable range, the method suffers from persistent jitter throughout the tracking process. This jitter originates from the numerical instability of the optimization-based trajectory planner, where small perturbations in target motion or constraints cause the solver to converge to different local optima at each planning cycle. Such jitter severely degrades visual quality and is particularly detrimental for gimbal-based compensation. Furthermore, the vertical error $e_V$ presents a systematic bias of approximately 0.5, which is larger than that of other methods, since vertical centering is neglected in the optimization cost and high-level command lacks the ability to control pitch angle.

\textbf{Ours}: Our method demonstrates the highest stability among all algorithms, avoiding the jitter observed in Elastic through continuous inference and the oscillations in PPO by leveraging first-order gradients for training. In terms of accuracy, it achieves performance comparable to the best baselines, with only minor deviations. Although it may appear slightly worse than PPO after convergence or Elastic in terms of $e_H$ at $v=1.5$\,m/s for Trajectories~D and~8, the target consistently remains within the most centralized 10\% region of the image plane, which fully meets the practical requirements of perception-aware tracking.

\subsection{Perspective Generalization}
\begin{figure}[h]
    \centering
    \includegraphics[width=1.0\linewidth]{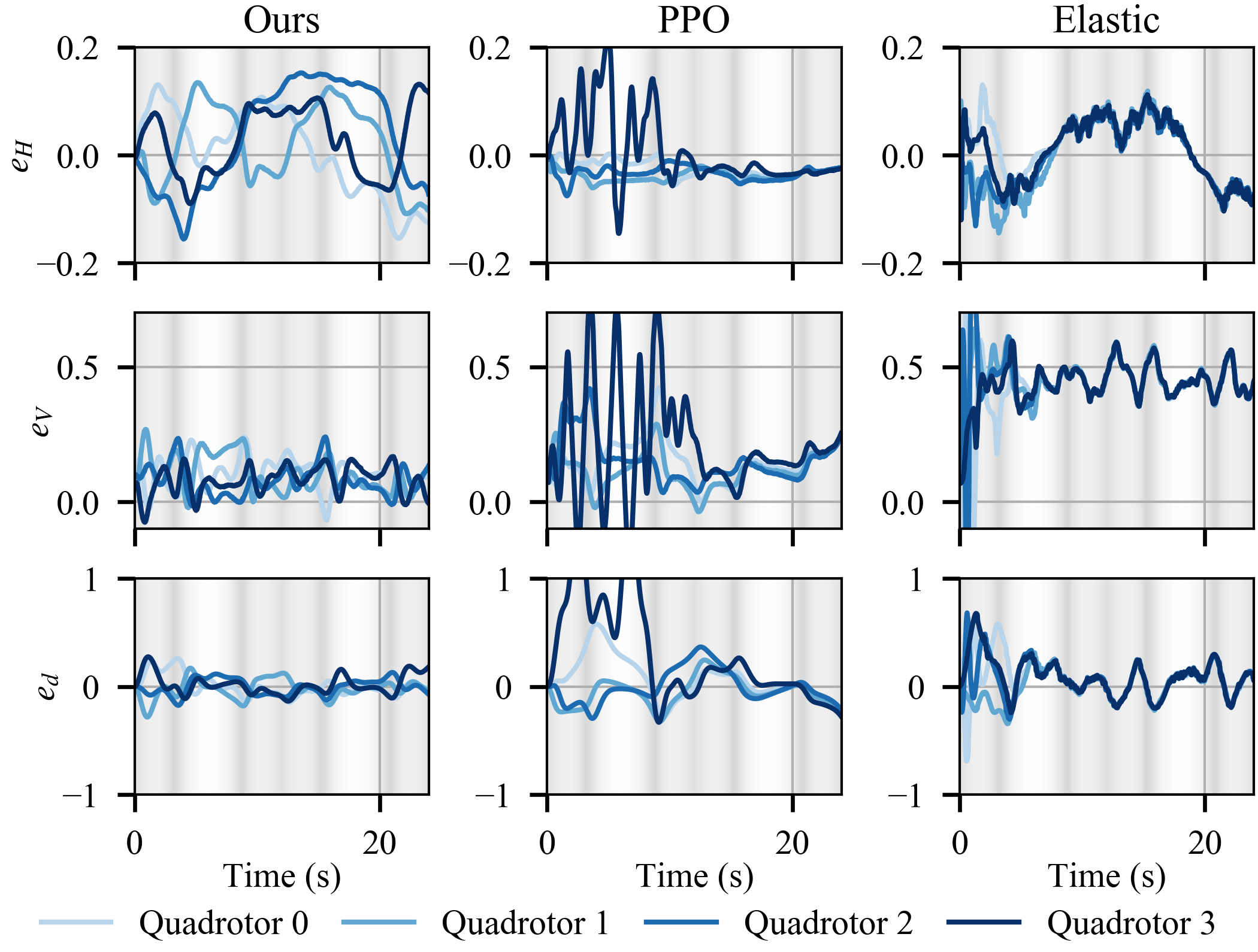}
    \caption{Error curves of four quadrotors tracking a target moving along Trajectory~8 with a velocity of 1.5\,m/s. The quadrotors are initialized at four perspectives: front, left, rear, and right.}
    \label{fig:exp_persp}
\end{figure}

Current studies have only examined tracking performance from the rear perspective. For more comprehensive analysis, four quadrotors are initialized from four different positions, as shown in \reffig{fig:objTrack}, and their performance is illustrated in \reffig{fig:exp_persp}.
The initial perspective has a strong influence on the tracking error along a trajectory. Elastic consistently suffers from an initial re-planning phase regardless of perspective.
PPO performs well for Quadrotors~0–2 but exhibits oscillations for Quadrotor~3, clearly indicating that it overfits to the perspectives of Quadrotors~0–2.  

In contrast, as revealed by the $e_H$ and $e_V$ curves, only our method adapts effectively to different initial perspectives over time, proving the generalization for various beginnings. Other algorithms eventually converge to a similar perspective, as reflected by their error curves gradually coinciding. This effect is particularly pronounced for Elastic, where almost all quadrotors converge to a state of following the target from behind after approximately 7 seconds (see supplementary video).  

\subsection{Overall Stability and Accuracy}

\begin{figure}[h]
    \centering
    \includegraphics[width=1.0\linewidth]{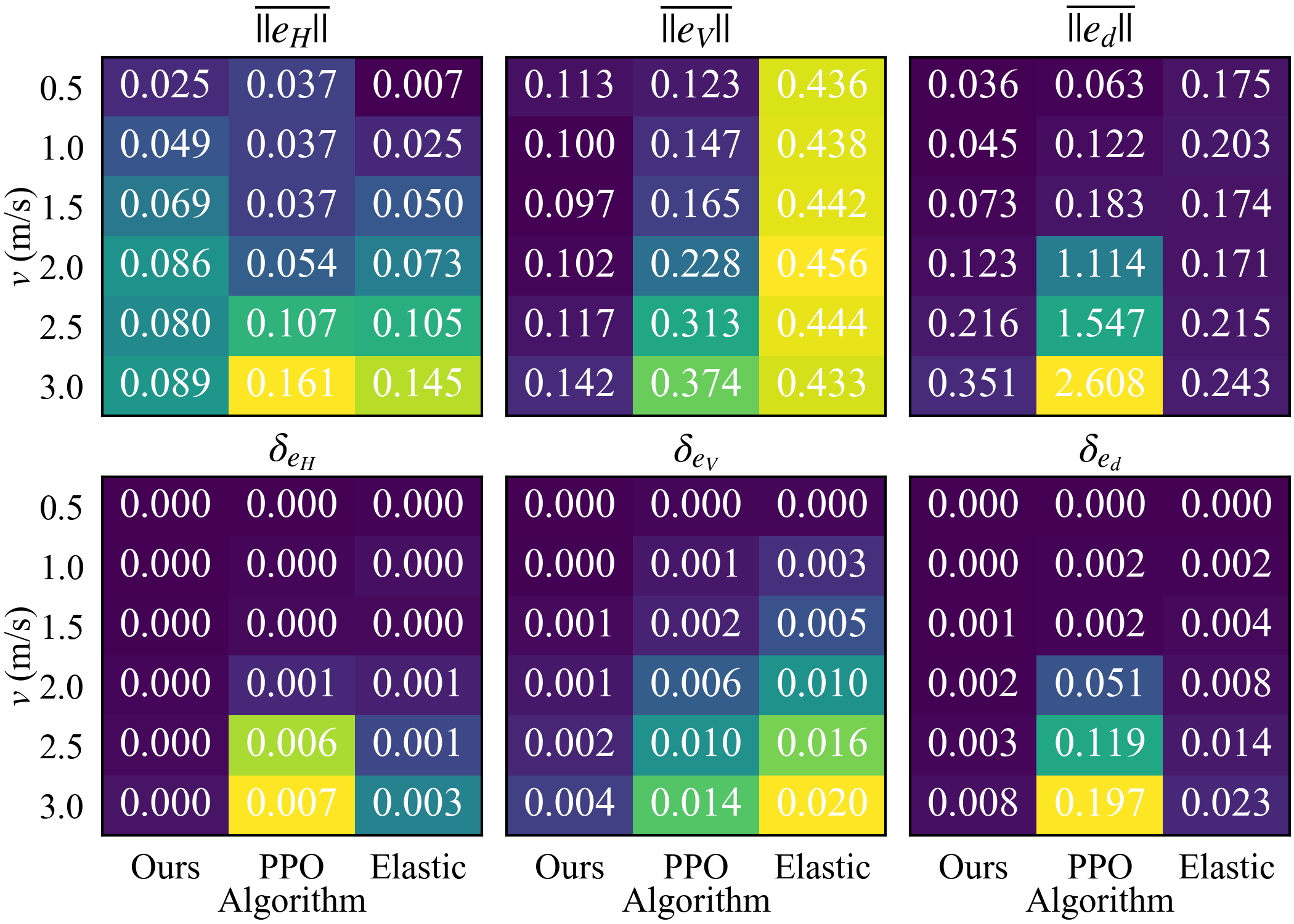}
    \caption{Mean absolute error and standard deviation across the tracking process of all three trajectories, tested with velocities ranging from 0.5\,m/s to 3.0\,m/s. The color intensity denotes the quality of each metric, with darker shades indicating better performance. The standard deviation is computed within a 0.5 second moving window for each quadrotor, providing a fair evaluation of short-term stability.}
    \label{fig:exp_overall_v}
\end{figure}

\reffig{fig:exp_overall_v} presents the mean absolute error and standard deviation across trajectories and quadrotors, which represent tracking accuracy and stability, respectively. Since stability characterizes the ability to remain consistent over short durations, $\delta$ at each moment is computed within a moving window. As expected, both error and deviation increase as the target velocity grows. PPO exhibits a pronounced degradation when the velocity exceeds 2.0\,m/s, as several quadrotors fail to maintain tracking; however, its performance remains comparable to ours at lower velocities below 2.0\,m/s. Elastic achieves accuracy ($e_H$, $e_d$) comparable to ours, but its short-term stability, reflected in $\delta_{e_V}$, $\delta_{e_H}$, and $\delta_{e_d}$, is at least twice worse than ours.  

\subsection{Distance Robustness}

\begin{figure}[h]
    \centering
    \includegraphics[width=1.0\linewidth]{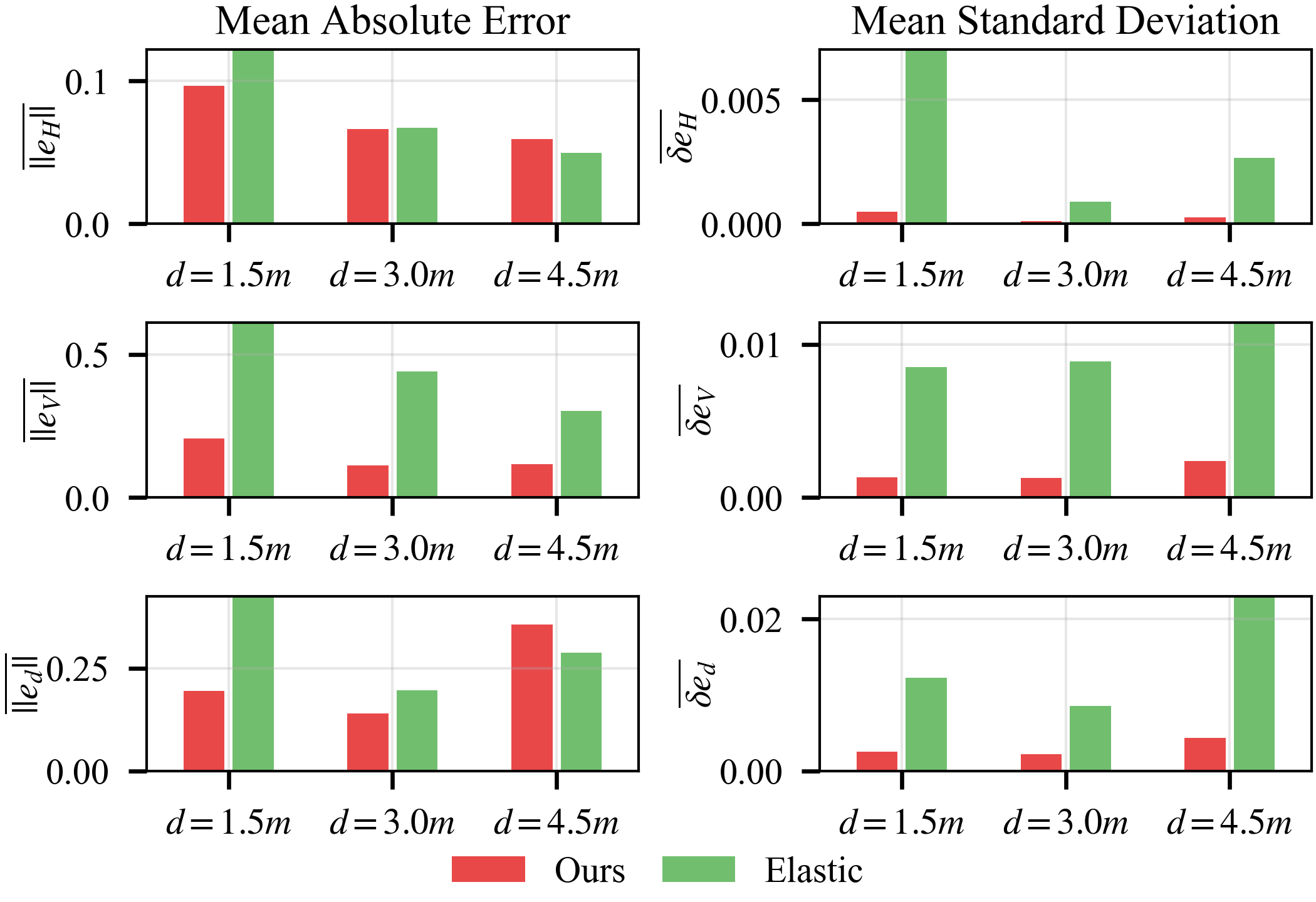}
    \caption{Tracking performance of our method and Elastic evaluated under three different distances: $1.5$\,m, $3.0$\,m, and $4.5$\,m. The left column shows the mean absolute error averaged over three trajectories and six velocities, while the right column reports the corresponding mean standard deviation.}
    \label{fig:exp_dis}
\end{figure}

We further evaluate policy performance across a range of safe distances from $1.5$\,m to $4.5$\,m. Since PPO diverges at higher velocities, its results are excluded from this analysis. As shown in \reffig{fig:exp_dis}, the results reveal a general trend: perception errors $e_V$ and $e_H$ decrease as the tracking distance increases, primarily due to reduced motion parallax. Meanwhile, the deviation of our method’s perception error grows only marginally and can be considered negligible.  

Overall, 3.0\,m emerges as the most appropriate tracking distance, achieving the lowest mean error and standard deviation while maintaining a smooth error variation pattern that is easier to compensate for during post-processing. More specifically, although performance at both $1.5$\,m and $4.5$\,m is inferior to that at 3.0\,m, the underlying causes differ. At $1.5$\,m, the quadrotor has limited maneuvering space to adjust its attitude when the target executes sharp turns. In contrast, at $4.5$\,m, the larger separation reduces sensitivity to distance variations, resulting in less responsive tracking.

\subsection{Real World Experiments}
We conducted a real-world one-take experiment to validate our policy, evaluating its performance in tracking a target undergoing continuous maneuvers. A DJI quadrotor was selected as the moving target due to its maneuverability, \lia{whose odometry is provided by external motion capture system, because we prioritizes the advancement of a better attitude-control algorithm for tracking}. The target executed a sequence of common maneuvers, while our FPV quadrotor driven by StableTracker tracked it. \reffig{fig:exp_real_world} illustrates the overall trajectory and target motion over time. The trajectory can be divided into the following stages: hover, acceleration to 5 m/s, deceleration to 1.5 m/s, ascent, turn-around, descent, S-shaped maneuver, and stop. 
In the supplementary video, the target remains steadily centered in the field of view throughout the entire trajectory, regardless of the maneuver. Moreover, the quadrotor exhibits the learned coordinated ascent–descent maneuver when encountering rapid acceleration and deceleration of target, as discussed in \refsec{sec:acc_dec}. A more detailed account of the tracking process is provided in the supplementary video.

\section{Discussion}
We presented a training solution that effectively mitigates computational overload and substantially improves learning-based policies compared with previous approaches. To focus on the core challenge of tracking, we simplified the problem by neglecting the target recognition module and instead providing ground-truth target states. Nevertheless, we have already developed a prototype object detector and pose estimator, which will be integrated in future.  

To further enhance applicability, we aim to train a policy capable of not only maintaining precise target tracking but also ensuring collision-free flight. While obstacle avoidance techniques are mature, combining it with high-accuracy tracking and stable flight control remains a challenging problem that will be addressed in our future work.

\section{Conclusion}

In this paper, we proposed \textbf{StableTracker}, an end-to-end policy predicting low-level commands for \textbf{stably} target tracking that maintains the target at the center of the image while preserving a fixed relative distance. We employed backpropagation-through-time (BPTT) via differentiable simulation to train the policy. A tailored reward structure was also introduced to facilitate policy training in tracking tasks. Comprehensive experiments against state-of-the-art traditional tracking methods and model-free reinforcement learning algorithms demonstrated that StableTracker achieves superior generalization, accuracy and stability. In particular, the BPTT-trained policy exhibited strong robustness across varying velocities, safe distances, and target trajectories. 
The FPV quadrotor also learns a novel maneuver—coordinated ascent and descent—that avoids losing the target under rapid acceleration or deceleration.
Finally, real-world onboard experiments in which a quadrotor tracked another quadrotor performing common maneuvers, confirmed the practicality of our approach.

\addtolength{\textheight}{-12cm}   

\bibliographystyle{IEEEtran}
\bibliography{references.bib}

@misc{lu_yopov2-tracker_2025,
	title = {{YOPOv2}-{Tracker}: {An} {End}-to-{End} {Agile} {Tracking} and {Navigation} {Framework} from {Perception} to {Action}},
	shorttitle = {{YOPOv2}-{Tracker}},
	doi = {10.48550/arXiv.2505.06923},
	urldate = {2025-07-09},
	publisher = {arXiv},
	author = {Lu, Junjie and Hui, Yulin and Zhang, Xuewei and Feng, Wencan and Shen, Hongming and Li, Zhiyu and Tian, Bailing},
	month = may,
	year = {2025},
}

@article{dionigi_d-vat_2024,
	title = {D-{VAT}: {End}-to-{End} {Visual} {Active} {Tracking} for {Micro} {Aerial} {Vehicles}},
	volume = {9},
	issn = {2377-3766},
	shorttitle = {D-{VAT}},
	doi = {10.1109/LRA.2024.3385700},
	number = {6},
	urldate = {2025-07-09},
	journal = {IEEE Robotics and Automation Letters},
	author = {Dionigi, Alberto and Felicioni, Simone and Leomanni, Mirko and Costante, Gabriele},
	month = jun,
	year = {2024},
	pages = {5046--5053},
}

@inproceedings{sampedro_image-based_2018,
	title = {Image-{Based} {Visual} {Servoing} {Controller} for {Multirotor} {Aerial} {Robots} {Using} {Deep} {Reinforcement} {Learning}},
	doi = {10.1109/IROS.2018.8594249},
	urldate = {2025-07-12},
	booktitle = {2018 {IEEE}/{RSJ} {International} {Conference} on {Intelligent} {Robots} and {Systems} ({IROS})},
	author = {Sampedro, Carlos and Rodriguez-Ramos, Alejandro and Gil, Ignacio and Mejias, Luis and Campoy, Pascual},
	month = oct,
	year = {2018},
	pages = {979--986},
}

@inproceedings{ji_elastic_2022,
	title = {Elastic {Tracker}: {A} {Spatio}-temporal {Trajectory} {Planner} for {Flexible} {Aerial} {Tracking}},
	shorttitle = {Elastic {Tracker}},
	doi = {10.1109/ICRA46639.2022.9811688},
	urldate = {2025-08-03},
	booktitle = {2022 {International} {Conference} on {Robotics} and {Automation} ({ICRA})},
	author = {Ji, Jialin and Pan, Neng and Xu, Chao and Gao, Fei},
	month = may,
	year = {2022},
	pages = {47--53},
}

@inproceedings{wang_visibility-aware_2021,
	title = {Visibility-aware {Trajectory} {Optimization} with {Application} to {Aerial} {Tracking}},
	doi = {10.1109/IROS51168.2021.9636753},
	urldate = {2025-08-03},
	booktitle = {2021 {IEEE}/{RSJ} {International} {Conference} on {Intelligent} {Robots} and {Systems} ({IROS})},
	author = {Wang, Qianhao and Gao, Yuman and Ji, Jialin and Xu, Chao and Gao, Fei},
	month = sep,
	year = {2021},
	note = {ISSN: 2153-0866},
	pages = {5249--5256},
}

@article{gao_adaptive_2024,
	title = {Adaptive {Tracking} and {Perching} for {Quadrotor} in {Dynamic} {Scenarios}},
	volume = {40},
	issn = {1552-3098, 1941-0468},
	doi = {10.1109/TRO.2023.3335670},
	urldate = {2025-08-03},
	journal = {IEEE Transactions on Robotics},
	author = {Gao, Yuman and Ji, Jialin and Wang, Qianhao and Jin, Rui and Lin, Yi and Shang, Zhimeng and Cao, Yanjun and Shen, Shaojie and Xu, Chao and Gao, Fei},
	year = {2024},
	note = {arXiv:2312.11866 [cs]},
	pages = {499--519},
}

@article{wang_image-based_2023,
	title = {Image-{Based} {Visual} {Servoing} of {Quadrotors} to {Arbitrary} {Flight} {Targets}},
	volume = {8},
	issn = {2377-3766},
	doi = {10.1109/LRA.2023.3245416},
	number = {4},
	urldate = {2025-08-03},
	journal = {IEEE Robotics and Automation Letters},
	author = {Wang, Guojie and Qin, Jiahu and Liu, Qingchen and Ma, Qichao and Zhang, Cong},
	month = apr,
	year = {2023},
	pages = {2022--2029},
}

@article{xi_anti-distractor_2022,
	title = {Anti-{Distractor} {Active} {Object} {Tracking} in {3D} {Environments}},
	volume = {32},
	issn = {1558-2205},
	doi = {10.1109/TCSVT.2021.3107153},
	number = {6},
	urldate = {2025-08-03},
	journal = {IEEE Transactions on Circuits and Systems for Video Technology},
	author = {Xi, Mao and Zhou, Yun and Chen, Zheng and Zhou, Wengang and Li, Houqiang},
	month = jun,
	year = {2022},
	pages = {3697--3707},
}

@article{qin_perception-aware_2023,
	title = {Perception-{Aware} {Image}-{Based} {Visual} {Servoing} of {Aggressive} {Quadrotor} {UAVs}},
	volume = {28},
	issn = {1941-014X},
	doi = {10.1109/TMECH.2023.3276211},
	number = {4},
	urldate = {2025-08-03},
	journal = {IEEE/ASME Transactions on Mechatronics},
	author = {Qin, Chao and Yu, Qiuyu and Go, H. S. Helson and Liu, Hugh H.-T.},
	month = aug,
	year = {2023},
	pages = {2020--2028},
}

@inproceedings{chen_tracking_2016,
	title = {Tracking a moving target in cluttered environments using a quadrotor},
	doi = {10.1109/IROS.2016.7759092},
	urldate = {2025-08-05},
	booktitle = {2016 {IEEE}/{RSJ} {International} {Conference} on {Intelligent} {Robots} and {Systems} ({IROS})},
	author = {Chen, Jing and Liu, Tianbo and Shen, Shaojie},
	month = oct,
	year = {2016},
	note = {ISSN: 2153-0866},
	pages = {446--453},
}

@article{zhao_hierarchical_2021,
	title = {Hierarchical {Active} {Tracking} {Control} for {UAVs} via {Deep} {Reinforcement} {Learning}},
	volume = {11},
	copyright = {http://creativecommons.org/licenses/by/3.0/},
	issn = {2076-3417},
	doi = {10.3390/app112210595},
	language = {en},
	number = {22},
	urldate = {2025-08-19},
	journal = {Applied Sciences},
	author = {Zhao, Wenlong and Meng, Zhijun and Wang, Kaipeng and Zhang, Jiahui and Lu, Shaoze},
	month = jan,
	year = {2021},
	pages = {10595},
}

@incollection{mozer2013focused,
  title={A focused backpropagation algorithm for temporal pattern recognition},
  author={Mozer, Michael C},
  booktitle={Backpropagation},
  pages={137--169},
  year={2013},
  publisher={Psychology Press}
}

@article{li2024visfly,
  title={Visfly: An efficient and versatile simulator for training vision-based flight},
  author={Li, Fanxing and Sun, Fangyu and Zhang, Tianbao and Zou, Danping},
  journal={arXiv preprint arXiv:2407.14783},
  year={2024}
}

@article{zhang_back_2024,
  title={Back to Newton's Laws: Learning Vision-based Agile Flight via Differentiable Physics},
  author={Zhang, Yuang and Hu, Yu and Song, Yunlong and Zou, Danping and Lin, Weiyao},
  journal={arXiv preprint arXiv:2407.10648},
  year={2024}
}

@article{loquercio_learning_2021,
	title = {Learning high-speed flight in the wild},
	volume = {6},
	issn = {2470-9476},
	pages = {eabg5810},
        year = {2021},
	number = {59},
	journal = {Science Robotics},
	author = {Loquercio, Antonio and Kaufmann, Elia and Ranftl, René and Müller, Matthias and Koltun, Vladlen and Scaramuzza, Davide},
	date = {2021},
}

@article{li2025abpt,
  title={ABPT: Amended Backpropagation through Time with Partially Differentiable Rewards},
  author={Li, Fanxing and Sun, Fangyu and Zhang, Tianbao and Zou, Danping},
  journal={arXiv preprint arXiv:2501.14513},
  year={2025}
}

@article{schulman2017proximal,
  title={Proximal policy optimization algorithms},
  author={Schulman, John and Wolski, Filip and Dhariwal, Prafulla and Radford, Alec and Klimov, Oleg},
  journal={arXiv preprint arXiv:1707.06347},
  year={2017}
}

@inproceedings{haarnoja2018soft,
  title={Soft actor-critic: Off-policy maximum entropy deep reinforcement learning with a stochastic actor},
  author={Haarnoja, Tuomas and Zhou, Aurick and Abbeel, Pieter and Levine, Sergey},
  booktitle={International conference on machine learning},
  pages={1861--1870},
  year={2018},
  organization={Pmlr}
}

@article{xu2022accelerated,
  title={Accelerated policy learning with parallel differentiable simulation},
  author={Xu, Jie and Makoviychuk, Viktor and Narang, Yashraj and Ramos, Fabio and Matusik, Wojciech and Garg, Animesh and Macklin, Miles},
  journal={arXiv preprint arXiv:2204.07137},
  year={2022}
}

@inproceedings{han_fast-tracker_2021,
	title = {Fast-{Tracker}: {A} {Robust} {Aerial} {System} for {Tracking} {Agile} {Target} in {Cluttered} {Environments}},
	shorttitle = {Fast-{Tracker}},
	doi = {10.1109/ICRA48506.2021.9561948},
	urldate = {2025-09-07},
	booktitle = {2021 {IEEE} {International} {Conference} on {Robotics} and {Automation} ({ICRA})},
	author = {Han, Zhichao and Zhang, Ruibin and Pan, Neng and Xu, Chao and Gao, Fei},
	month = may,
	year = {2021},
	note = {ISSN: 2577-087X},
	pages = {328--334},
}

@article{pan_fast-tracker_2021,
	title = {Fast-{Tracker} 2.0: {Improving} autonomy of aerial tracking with active vision and human location regression},
	volume = {3},
	copyright = {© 2021 The Authors. IET Cyber-Systems and Robotics published by John Wiley \& Sons Ltd on behalf of Zhejiang University Press.},
	issn = {2631-6315},
	shorttitle = {Fast-{Tracker} 2.0},
	doi = {10.1049/csy2.12033},
	language = {en},
	number = {4},
	urldate = {2025-09-07},
	journal = {IET Cyber-Systems and Robotics},
	author = {Pan, Neng and Zhang, Ruibin and Yang, Tiankai and Cui, Can and Xu, Chao and Gao, Fei},
	year = {2021},
	pages = {292--301},
}

@article{zhou2020ego,
  title={Ego-planner: An esdf-free gradient-based local planner for quadrotors},
  author={Zhou, Xin and Wang, Zhepei and Ye, Hongkai and Xu, Chao and Gao, Fei},
  journal={IEEE Robotics and Automation Letters},
  volume={6},
  number={2},
  pages={478--485},
  year={2020},
  publisher={IEEE}
}


\end{document}